%% file: main_ieee.tex
\documentclass[lettersize,journal]{IEEEtran}
\usepackage{amsmath,amsfonts}
\usepackage{algorithm}
\usepackage{array}
\usepackage[caption=false,font=normalsize,labelfont=sf,textfont=sf]{subfig}
\usepackage{textcomp}
\usepackage{stfloats}
\usepackage{url}
\usepackage{verbatim}
\usepackage{graphicx}
\usepackage{cite}
\input{commands.tex}

\begin{document}

\title{Hyperparameter Optimization for SecureBoost via Constrained Multi-Objective Federated Learning}





\author{\IEEEauthorblockN{Yan Kang$^{\dagger}$\IEEEauthorrefmark{1},
Ziyao Ren$^{\dagger}$,  
Lixin Fan,
Linghua Yang,
Yongxin Tong, 
and Qiang Yang}

\IEEEcompsocitemizethanks{\IEEEcompsocthanksitem Ziyao Ren, Linghua Yang, and Yongxin Tong are with the School of Computer Science, Beihang University, Beijing, China. 
\IEEEcompsocthanksitem Yan Kang, Lixin Fan, and Qiang Yang are with Webank, Shenzhen, China. Qiang Yang is also with Hong Kong University of Science and Technology. 
\IEEEcompsocthanksitem Yan Kang and Ziyao Ren are co-first authors.
}

\thanks{Corresponding author: Yan Kang (email: yangkang@webank.com).}
}

\maketitle

\begin{abstract}
SecureBoost is a tree-boosting algorithm that leverages homomorphic encryption (HE) to protect data privacy in vertical federated learning. SecureBoost and its variants have been widely adopted in fields such as finance and healthcare. However, the hyperparameters of SecureBoost are typically configured heuristically for optimizing model performance (i.e., utility) solely,  assuming that privacy is secured. Our study found that SecureBoost and some of its variants are still vulnerable to label leakage. This vulnerability may lead the current heuristic hyperparameter configuration of SecureBoost to a suboptimal trade-off between utility, privacy, and efficiency, which are pivotal elements toward a trustworthy federated learning system. To address this issue, we propose the Constrained Multi-Objective SecureBoost (CMOSB) algorithm, which aims to approximate Pareto optimal solutions that each solution is a set of hyperparameters achieving an optimal trade-off between utility loss, training cost, and privacy leakage. We design measurements of the three objectives, including a novel label inference attack named instance clustering attack (ICA) to measure the privacy leakage of SecureBoost. Additionally, we provide two countermeasures against ICA. The experimental results demonstrate that the CMOSB yields superior hyperparameters over those optimized by grid search and Bayesian optimization regarding the trade-off between utility loss, training cost, and privacy leakage.
\end{abstract}


\input{sections/1_intro}
\input{sections/2_related}
\input{sections/3_background}
\input{sections/4_method}
\input{sections/5_exp}
\input{sections/6_conclusion}


\bibliographystyle{IEEEtran}
\bibliography{ijcai23}


 




\vfill

\end{document}

%% file: commands.tex
\usepackage{afterpage}
\usepackage{makecell}
\usepackage{tabulary}
\usepackage{xcolor}
\usepackage{colortbl}
\usepackage{prettyref}
\usepackage{framed}
\usepackage{booktabs}       

\newcommand{\savehyperref}[2]{\texorpdfstring{\hyperref[#1]{#2}}{#2}}
\newrefformat{eq}{\savehyperref{#1}{Eq. \textup{(\ref*{#1})}}}
\newrefformat{eqn}{\savehyperref{#1}{Eq.~\textup{(\ref*{#1})}}}
\newrefformat{lem}{\savehyperref{#1}{Lemma~\ref*{#1}}}
\newrefformat{assump}{\savehyperref{#1}{Assumption~\ref*{#1}}}
\newrefformat{defi}{\savehyperref{#1}{Definition~\ref*{#1}}}\newrefformat{tab}{\savehyperref{#1}{Table~\ref*{#1}}}
\newrefformat{lemma}{\savehyperref{#1}{Lemma~\ref*{#1}}}
\newrefformat{line}{\savehyperref{#1}{Line~\ref*{#1}}}
\newrefformat{thm}{\savehyperref{#1}{Theorem~\ref*{#1}}}
\newrefformat{corr}{\savehyperref{#1}{Corollary~\ref*{#1}}}
\newrefformat{cor}{\savehyperref{#1}{Corollary~\ref*{#1}}}
\newrefformat{sec}{\savehyperref{#1}{Section~\ref*{#1}}}
\newrefformat{app}{\savehyperref{#1}{Appendix~\ref*{#1}}}
\newrefformat{ex}{\savehyperref{#1}{Example~\ref*{#1}}}
\newrefformat{fig}{\savehyperref{#1}{Figure~\ref*{#1}}}
\newrefformat{alg}{\savehyperref{#1}{Algorithm~\ref*{#1}}}
\newrefformat{rem}{\savehyperref{#1}{Remark~\ref*{#1}}}
\newrefformat{conj}{\savehyperref{#1}{Conjecture~\ref*{#1}}}
\newrefformat{prop}{\savehyperref{#1}{Proposition~\ref*{#1}}}
\newrefformat{proto}{\savehyperref{#1}{Protocol~\ref*{#1}}}
\newrefformat{prob}{\savehyperref{#1}{Problem~\ref*{#1}}}
\newrefformat{claim}{\savehyperref{#1}{Claim~\ref*{#1}}}
\newrefformat{que}{\savehyperref{#1}{Question~\ref*{#1}}}
\newrefformat{op}{\savehyperref{#1}{Open Problem~\ref*{#1}}}
\newrefformat{fn}{\savehyperref{#1}{Footnote~\ref*{#1}}}
\newrefformat{property}{\savehyperref{#1}{Property~\ref*{#1}}}

\usepackage{xspace}
\usepackage{amsmath}
\usepackage{amsthm}
\usepackage{bbm}
\usepackage{algorithm}
\usepackage[noend]{algpseudocode}
\usepackage{mathrsfs}
\usepackage{amssymb}
\usepackage{bm}
\usepackage{makecell}
\usepackage{tabulary}

\newcommand{\blue}[1]{{\color{blue}#1}}

\newcommand{\gray}[1]{{\color{gray}#1}}

\newtheorem{definition}{Definition}

\newtheorem{remark}{Remark}

\theoremstyle{definition}

\newcommand{\fakeparagraph}[1]{\vspace{1mm}\noindent\textbf{#1.}}

\newif\ifrevflag
\revflagfalse

\newcommand{\setrevflag}[1]{%
  \ifnum#1=0
    \revflagfalse
  \else
    \revflagtrue
  \fi
}

\newcommand{\rev}[1]{%
  \ifrevflag
    \blue{#1}
  \else
    #1
  \fi
}

%% file: sections/1_intro.tex
\section{Introduction}

Federated learning (FL)~\cite{DBLP:conf/aistats/McMahanMRHA17} is a novel distributed machine learning paradigm that enables multiple participants to train machine learning models without compromising data privacy. 
Federated learning can be categorized into horizontal federated learning and vertical federated learning based on how the data is distributed among participating parties~\cite{DBLP:journals/tist/YangLCT19}. Vertical federated learning (VFL)~\cite{liu2022vertical} refers to the scenario where feature data is vertically partitioned among multiple parties, with one party possessing the data labels. In VFL, each party holds different parts of the feature data, and they collaborate to train machine learning models without directly sharing their raw data. SecureBoost~\cite{DBLP:journals/expert/ChengFJLCPY21} is a widely adopted vertical federated tree-boosting algorithm for its interpretability and privacy-preserving ability. Nevertheless, It has the following two limitations.

\textbf{Limitation 1: SecureBoost still faces the possibility of label leakage~\cite{DBLP:conf/uss/Fu0JCWG0L022} through intermediate information, despite employing homomorphic encryption to protect instance gradients.} 
Our experimental results in Sec.~\ref{sec:exp-def} reveal that up to 84\% of labels can be leaked stemming from the absence of protection on instance distributions. Therefore, defense mechanisms are required for SecureBoost to protect the label privacy in addition to HE.


\textbf{Limitation 2: Heuristic hyperparameter configuration may lead to suboptimal trade-off between utility, efficiency, and privacy of the SecureBoost model.} Existing FL platforms~\cite{DBLP:journals/jmlr/LiuFCXY21,DBLP:conf/sigmod/FuSYJXT021} typically determine the hyperparameters of SecureBoost heuristically, which may lead to suboptimal hyperparameter choices that do not maximize utility, efficiency, and privacy - the pivotal elements of trustworthy federated learning.

To address the two limitations, we propose Constrained Multi-Objective SecureBoost (CMOSB)~\cite{kang2023optimizing} to find Pareto optimal solutions of hyperparameters that can simultaneously minimize three conflicting objectives: \textit{utility loss}, \textit{training cost}, and \textit{privacy leakage}. Each solution represents an optimal trade-off between the three objectives~\cite{zhang2023tradeoff}. Consequently, Pareto optimal solutions not only provide optimal hyperparameters for SecureBoost but also cater to the flexible requirements of VFL participants concerning privacy and resource constraints. For example, participants can select the most appropriate hyperparameters from the Pareto optimal solutions that align with their preference for utility, efficiency, and privacy. Our main contributions are summarized as follows: 
\begin{itemize}
    \item We formalize the Constrained Multi-Objective SecureBoost (CMOSB) problem and correspondingly propose a CMOSB algorithm to identify Pareto optimal solutions of hyperparameters that simultaneously minimize utility loss, training cost, and privacy leakage. Our CMOSB algorithm can be readily extended to optimize other objectives, providing a versatile and effective approach to addressing the complex trade-offs in training federated tree-boosting models.
    \item We design measurements of utility loss, training cost, and privacy leakage. In particular, we propose a novel label inference attack named instance clustering attack to measure privacy leakage. We also develop two countermeasures against this attack.
    \item We conduct experiments on four datasets, demonstrating that our CMOSB can find better hyperparameters than grid search and Bayesian optimization in terms of the trade-off between privacy leakage, utility loss, and training cost. Moreover, CMOSB can find Pareto optimal solutions of hyperparameters that achieve better trade-offs between utility loss, training cost, and privacy leakage than state-of-the-art solutions.
    
\end{itemize}

The rest of the paper is organized as follows. We first review related work in Sec.~\ref{sec:rel} and introduce the preliminary in Sec.~\ref{sec:bac}. Then, we describe the CMOSB problem in Sec.~\ref{sec:cmosb_problem} and elaborate on our proposed label inference attack and the corresponding defense methods in Sec.~\ref{sec:priv}. Next, we provide the CMOSB algorithm in Sec.~\ref{sec:cmosb}. We report our experimental results in Sec.~\ref{sec:exp} and conclude this paper in Sec.~\ref{sec:conclusion}.

%% file: sections/2_related.tex
\section{Related Work}\label{sec:rel}

In this section, we review related work from three categories: tree-based models in VFL, label leakage in VFL, and multi-objective federated learning. 

\subsection{Tree-based Models in VFL}

Tree-based models are a common type of machine learning algorithm, and their applications in federated learning require the design of appropriate privacy protection methods. 
SecureBoost~\cite{DBLP:journals/expert/ChengFJLCPY21} is an XGBoost-based model that protects data privacy by applying homomorphic encryption to intermediate gradients.
SecureBoost+~\cite{Chen2021secureboostplus} extends the SecureBoost algorithm to multi-class tasks and improves SecureBoost's training efficiency. 
VF\textsuperscript{2}Boost~\cite{DBLP:conf/sigmod/FuSYJXT021} is a system based on SecureBoost that reduces model training costs through parallel computation.
HEP-XGB~\cite{DBLP:conf/cikm/FangZT0YWWZZ21} employs a customized secret sharing method to achieve efficient two-party XGBoost model construction.
Pivot~\cite{DBLP:journals/pvldb/WuCXCO20} utilizes a combination of homomorphic encryption and secure multi-party computation to train XGBoost models, ensuring that no intermediate results are leaked during the training process.

\subsection{Label Leakage in VFL}


Label leakage in vertical federated learning refers to the situation where the passive party is able to obtain the label information of the active party during the training process. 
Fu et al.~\cite{DBLP:conf/uss/Fu0JCWG0L022} systematically classified the label leakage in vertical federated learning into three categories: active attack, passive attack, and direct attack, and provided an attack method for each category. 
Zou et al.~\cite{liu2021batch} proposed a label attack method on the black-boxed VFL and presented a privacy protection method.
Xie et al.~\cite{DBLP:journals/corr/abs-2301-07284} demonstrated that split learning remains vulnerable to label inference attacks.
Qiu et al.~\cite{LabelRelation22} found that the training of GNNs in VFL may also lead to the leakage of sample relationships.
Tan et al.~\cite{DBLP:conf/bigcom/TanZLLW22} proposed a gradient inversion attack that utilizes the gradients of local models to reconstruct label data.
Pan et al.~\cite{DBLP:journals/pvldb/PanTXZDZSZCXXL22} implements an efficient data federation system using secure multi-party computation to avoid label leakage.

\subsection{Multi-objective Federated Learning}

Multi-objective federated learning is a collaborative optimization approach in which participants of FL aim to optimize multiple conflicting objectives simultaneously and find Pareto optimal solutions.
Personalized Federated Learning aims to optimize the federated model structure for each participating party's data distribution\cite{DBLP:conf/icde/WangTZZPF023,zhang2023dm}.
Cui et al.~\cite{DBLP:conf/nips/CuiPLZW21} considered the utility of participating parties as the optimization objective, model disparity as the constraint, and optimized the objectives of all participants to calculate the Pareto front. This approach ensures the fairness of federated learning and enables obtaining the optimal performance model. 
Zhu et al.~\cite{DBLP:journals/tnn/ZhuJ20} formulated the accuracy and communication cost of federated learning as the optimization objectives and adjusted the model sparsity to minimize both the communication cost and test errors.
The algorithm proposed in \cite{kang2023optimizing} is a multi-objective federated learning algorithm with constraints, which adds constraints to the optimization objectives and finds the Pareto front more efficiently within the feasible range.

%% file: sections/3_background.tex
\section{Preliminary}
\label{sec:bac}

In this section, we introduce some preliminaries of vertical federated learning (VFL) and SecureBoost. 

\subsection{Vertical Federated Learning}

\begin{figure}[!h]
    \centering
    \includegraphics[width=0.99\linewidth]{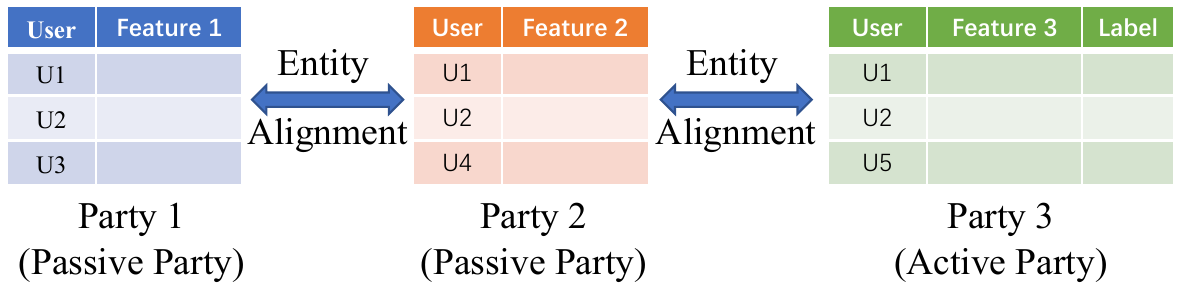}
    \caption{An illustration of data partition in VFL. }
    \label{fig:verticalfed}
\end{figure}

Vertical Federated Learning~\cite{DBLP:journals/tist/YangLCT19,liu2022vertical} is one of the scenarios in Federated Learning. In this setting, feature data is vertically partitioned among multiple parties, with one party possessing the data labels (Fig.~\ref{fig:verticalfed}). Specifically, the active party holds both features and labels, while the passive parties hold only features.

All participating parties in VFL first align their instances by private set intersection (PSI)~\cite{DBLP:conf/eurocrypt/FreedmanNP04}, and then perform federated model training and inference. 
During the training process, each party is not allowed to reveal its training data to others~\cite{DBLP:journals/pvldb/TongPZSXZZCXXL22}. 
After the federated training is completed, the federated model $M$ is obtained, which is jointly held by all participating parties, i.e., model $M$ is split into $M_1, M_2, \dots, M_K$. 
In the inference phase, all participating parties collaboratively make the prediction, but only the active party can access the final predicted result. 

\subsection{SecureBoost}
SecureBoost~\cite{DBLP:journals/expert/ChengFJLCPY21} is a widely used gradient boosting tree algorithm designed for the vertical federated learning scenario. The core idea of SecureBoost is to use $n$ federated decision trees to fit instance labels.


For each iteration, SecureBoost constructs a new tree from the root node based on a node split finding algorithm, which is summarized in Algo.~\ref{alg:split}. In this process, the active party sends the homomorphically encrypted gradients $\left\langle g\right\rangle$, $\left\langle h\right\rangle$, and instance space $I$ of the current node to passive parties. The passive parties then calculate the gradient statistics and send them back to the active party. The active party finds the optimal split of the current node with the maximal splitting score and informs the corresponding parties to split the instance space into child nodes.
The split-finding process continues until the maximum depth is reached. For a more detailed description, please refer to \cite{DBLP:journals/expert/ChengFJLCPY21}.

\begin{algorithm}[!h]
    \caption{SecureBoost Split Finding (SF)}
	\begin{algorithmic}[1]
    \vspace{2pt}
    \Statex \textbf{Input:} Instance space of current node $I$; 
    \Statex \textbf{Input:} Gradient $\left\langle g\right\rangle$, hessian $\left\langle h\right\rangle$.
    \Statex \textbf{Output:} Partition current instance space according to the selected attribute's value. 
    \State \gray{$\triangleright$ \textit{Passive parties perform:}}


  \State {Calculate gradient statistics based on $\left\langle g\right\rangle$, $\left\langle h\right\rangle$, and $I$; }
  
    \State \gray{$\triangleright$ \textit{Active party performs:}}
    \For{each split point}
        \State {Calculate info gain based on gradient statistics;}
        \State {Update splitting score based on the info gain;}
    \EndFor
    \State {Return optimal split to the corresponding party;}
    \State \gray{$\triangleright$ \textit{Passive party performs:}}
    \State {Partition instance space to form $I_L$ according to the optimal split, and update model;}
    \State \gray{$\triangleright$ \textit{Active party performs:}}
    \State {Split current node according to $I_L$ and update model;}
	\end{algorithmic}\label{alg:split}
\end{algorithm}
The instance distributions of leaf nodes in SecureBoost are not protected, which may lead to privacy leakage. Some methods~\cite{DBLP:journals/pvldb/WuCXCO20,DBLP:conf/cikm/FangZT0YWWZZ21} use a combination of Homomorphic Encryption (HE) and Secure Multi-Party Computation (MPC) to address this issue and enhance privacy protection. However, these methods may result in significant training cost, making it challenging to apply them in practical and trustworthy federated learning scenarios.





\section{Constrained Multi-Objective SecureBoost Problem}\label{sec:cmosb_problem}
The Constrained Multi-Objective SecureBoost Learning (CMOSB) problem aims to find Pareto optimal solutions that each is a set of hyperparameters leading to an optimal trade-off between three objectives under constraints: utility loss, training cost, and privacy leakage. 
Inspired by \cite{kang2023optimizing}, we formulate the CMOSB problem as follows:
\begin{equation}\label{cmosb}
    \begin{split}
       &\min\limits_{x \in \mathcal{X}} ( \epsilon_{u}(x), \epsilon_{c}(x), \epsilon_{p}(x) ) \\
        & \text{subject to } \,\, \epsilon_{u}(x) \leq \phi_{u}, \epsilon_{c}(x) \leq \phi_{c}, \epsilon_{p}(x) \leq \phi_{p}.
    \end{split}
\end{equation}
where $x \in \mathbb{R}^d$ is a solution in the decision space $\mathcal{X}$; $\epsilon_{u}(x)$, $\epsilon_{c}(x)$, and $\epsilon_{p}(x)$ denote the objectives of utility loss, training cost, and privacy leakage, respectively; $\phi_u$, $\phi_c$, and $\phi_p$ are the upper bounds of $\epsilon_u$, $\epsilon_c$, and $\epsilon_p$, respectively.

\begin{remark}
In this paper, a solution $x$ refers to a collection of hyperparameters, such as the depth of decision trees, batch size, protection strength parameters, and so on. Each set of hyperparameters corresponds to specific values of privacy leakage, utility loss, and training cost. 
\end{remark}

During multi-objective SecureBoost training, SecureBoost participants put constraints on utility loss, training cost, and privacy leakage to ensure that the utility meets the specified requirements. These constraints also serve to limit training cost and privacy leakage within the allocated budget. By doing so, SecureBoost participants can strike a balance between utility, cost, and privacy in a controlled manner.

Formally, SecureBoost participants aim to find \textit{Pareto front} and corresponding \textit{Pareto optimal solutions} for the Constrained Multi-Objective SecureBoost problem. We provide the definitions of Pareto dominance, Pareto optimal solution, Pareto set, and Pareto front as follows.

\begin{definition}[Pareto Dominance]\label{def:pareto_dom}
Let $x_a, x_b \in \mathcal{X}$,  $x_a$ is said to dominate $x_b$, denoted as $x_a \prec x_b$, if and only if $f_i(x_a) \leq f_i(x_b), \forall i \in \{1,\ldots,m\}$ and $f_j(x_a) <  f_j(x_b), \exists j \in \{1,\ldots,m\}$.
\end{definition}

\begin{definition}[Pareto Optimal Solution]\label{def:pareto_sol}
A solution $x^{*} \in \mathcal{X}$ is called a Pareto optimal solution if there does not exist a solution $\hat{x} \in \mathcal{X}$ such that $\hat{x} \prec x^*$.
\end{definition}

A Pareto optimal solution refers to a solution that achieves an optimal trade-off among different conflicting objectives. The collection of all Pareto optimal solutions forms the Pareto set, while the corresponding objective values for the Pareto optimal solutions form the Pareto front. The definitions of the Pareto set and Pareto front are formally given as follows.


\begin{definition}[Pareto Set and Front]\label{def:pareto_set_front}
The set of all Pareto optimal solutions is called the Pareto set, and its image in the objective space is called the Pareto front.
\end{definition}


We use hypervolume~(HV) indicator~\cite{DBLP:conf/ppsn/ZitzlerK04} as a metric to measure the Pareto front. 
The definition of hypervolume indicator is provided below.

\begin{definition}[Hypervolume Indicator]\label{def:hypervolume}
Let $z = \{z_1,\cdots, z_m\}$ be a reference point that is an upper bound of the objectives $V = \{v_1,\ldots, v_m\}$, such that $v_{i} \leq z_i$, $\forall i \in [m]$. The hypervolume indicator $\text{HV}(V)$ measures the region between $V$ and $z$ and is formulated as:
\begin{equation}
    \text{HV}(V) = \Lambda \left( \left \{ q \in \mathbb{R}^m \big| q \in \prod_{i=1}^{m}[v_i, z_i]  \right \}\right) 
\end{equation}
where $\Lambda(\cdot)$ refers to the Lebesgue measure.
\end{definition}

To address the CMOSB problem outlined in Eq.(\ref{cmosb}), we must first quantify the utility loss, training cost, and privacy leakage of a given solution. Measuring utility loss and training cost is straightforward, which we will discuss in Section \ref{sec:objectives}. Evaluating privacy leakage requires a more sophisticated approach, as it involves assessing the extent to which an individual's private information can be inferred from the model's outputs. To address this challenge, we propose a novel label inference attack that exploits a vulnerability in SecureBoost, allowing us to infer private label information. In the next section, we will delve deeper into the threat model, the label inference attack, and potential countermeasures.

%% file: sections/4_method.tex
\section{Privacy Leakage in SecureBoost}
\label{sec:priv}

In this section, we first elaborate on the threat model. Then, we design a label inference attack and explain how this attack can lead to the leakage of labels owned by the active party. Finally, we propose two defense methods against this attack.

\subsection{Threat Model}
\label{sec:threat}
Below, we discuss the threat model, which includes the attacker's objective, capability, and knowledge.




\noindent\textbf{Attacker's objective}. We consider the passive party as the attacker, who aims to infer labels owned by the active party. 

\noindent\textbf{Attacker's capability}. We assume the attacker is \textit{semi-honest}, meaning that the attacker faithfully follows the SecureBoost protocol but may attempt to infer labels of the active party.

\noindent\textbf{Attacker's knowledge}. 
The passive party can access instance distributions of leaf nodes since SecureBoost does not protect this information. Additionally, we assume that the passive party holds several labeled instances for each class, similar to the assumption made in \cite{DBLP:conf/uss/Fu0JCWG0L022}.


\subsection{Label Inference Attack}
\label{sec:attack}

\rev{
Gradient boost decision tree (GBDT) construction involves computing information gain based on instance features owned by different parties. To ensure that private features are not disclosed to other parties during the process of calculating information gain, the active party needs to share instance gradients and distributions with passive parties in a privacy-preserving manner.
}
While SecureBoost and its variant~\cite{Chen2021secureboostplus} utilize homomorphic encryption to protect gradient information passed between the active party and passive parties, they do not protect the instance distribution of each leaf node owned by the passive party. This indicates that the passive party (i.e., the attacker) can exploit this information to infer the active party's labels through clustering.

\begin{figure}[!h]
    \centering
    \includegraphics[width=0.99\linewidth]{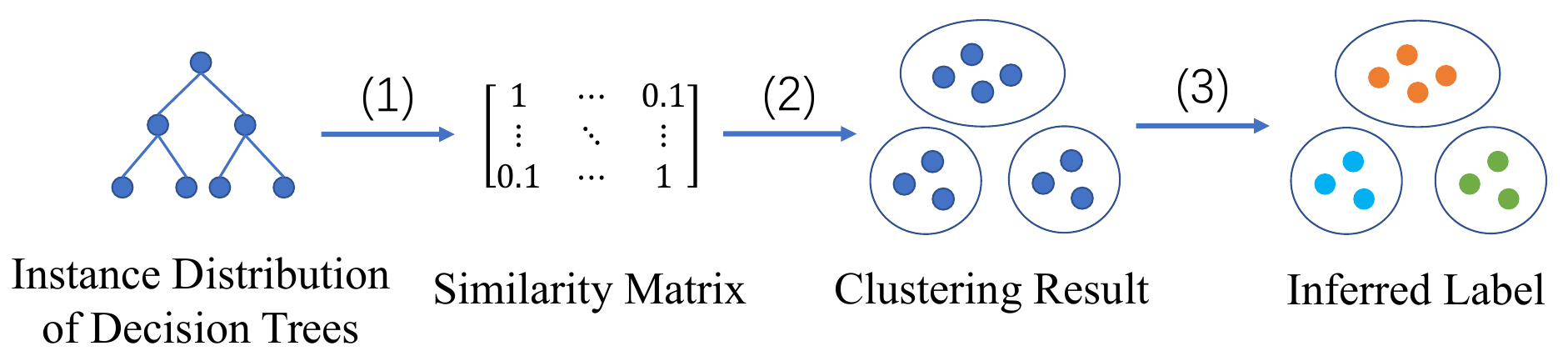}
    \caption{The workflow of instance clustering attack. 
    (1) The attacker constructs a similarity matrix based on the instance distribution.
    (2) The attacker clusters the training instances based on the similarity matrix.
    (3) The attacker infers labels of unlabeled instances based on the known labels.
    }
    \label{fig:attack}
\end{figure}

We propose a label inference attack named \textit{Instance Clustering Attack (ICA)} to infer the label information owned by the active party and leverage ICA to measure the privacy leakage of the SecureBoost algorithm. The procedure of ICA is illustrated in Fig.~\ref{fig:attack} and described in Algo.~\ref{alg:attack}. 

\begin{algorithm}[!h]
    \caption{Instance Clustering Attack}
	\begin{algorithmic}[1]
    \Statex \textbf{Input:} Instance distribution $\{I_i\}_{i=1}^{n}$. 
    \Statex \textbf{Output:} Predicted label $\hat{y}$. 
    \State {Calculate instance similarity matrix $S$ by Eq.(\ref{sim});}
    \State {Categorize instances into $C$ clusters $\{c_i\}_{i=1}^C$ based on $S$;}
    \For{$i=1$ to $C$}
        \State {$y_i \leftarrow$ the known label of one instance in cluster $c_i$; }
        \State {Assign label $y_i$ to $\hat{y}$ for all instances in $c_i$; }
    \EndFor
    \State \Return {$\hat{y}$;}
    \end{algorithmic}\label{alg:attack}
\end{algorithm}
 

The idea behind the attack is straightforward: samples with the same label are highly likely to be grouped into the same leaf node.
The attacker first constructs the similarity matrix based on the instance distribution of each leaf node using Eq.~(\ref{sim}) (line 1 of Algo.~\ref{alg:attack}).
\begin{equation}
\begin{split}
\label{sim}
sim(a, b)= & \frac{1}{n}\sum_{i=1}^{n}s(a, b, i), \\
\text{where  } s(a,b,i)= & \begin{cases}1,\text{ if } \exists j, a,b \in I_{i,j}\\ 0,\text{ otherwise} \end{cases}
\end{split}
\end{equation}
where $n$ denotes the number of decision trees, and $j$ denotes the index of the leaf node.

Then, the attacker utilizes this similarity matrix to categorize instances into $C$ clusters (line 2). We employ spectral clustering~\cite{DBLP:conf/nips/NgJW01}, a straightforward yet effective method for clustering instances based on the similarity matrix. This stage essentially involves learning a mapping from cluster IDs to the actual labels. If the attacker possesses more labeled data, it can establish a better mapping, which improves the accuracy of ICA. We assume the attacker knows the label of one instance in each cluster. Hence, the attacker assigns instances in each cluster with the known labels (lines 3-5). 

We use the accuracy of the instance clustering attack to measure the privacy leakage $\epsilon_p$ (see Sec. \ref{sec:objectives}).


\subsection{Defense Methods}
\label{sec:defense}


\rev{
Due to the high communication and computational overhead associated with directly preserving the privacy of the instance distribution, Secure Multi-Party Computation (MPC) is often challenging to apply in real-world scenarios. In this work, we propose two defense methods that can mitigate privacy leakage caused by the instance clustering attack (ICA).
}

The performance of ICA depends on the purity of leaf nodes owned by the attacker (i.e., the passive party). We define the purity of a node as the ratio of instances belonging to the majority class to all instances in that node. A higher purity implies a more accurate similarity matrix, which can lead to more precise clustering results and, consequently, more privacy leakage. Therefore, our proposed defense methods mitigate privacy leakage by reducing node purity, and we name the two defense methods:
(1) local trees
\footnote{
Our defense method has been implemented in FATE v1.11.2.
}; 
(2) purity threshold.



\fakeparagraph{Local Trees}

The mechanism of ICA (see Sec.~\ref{sec:attack}) indicates that the stronger the correlation between the instance distribution and true labels, the higher the likelihood of label privacy leakage.

\begin{figure}[!h]
    \centering \includegraphics[width=0.95\linewidth]{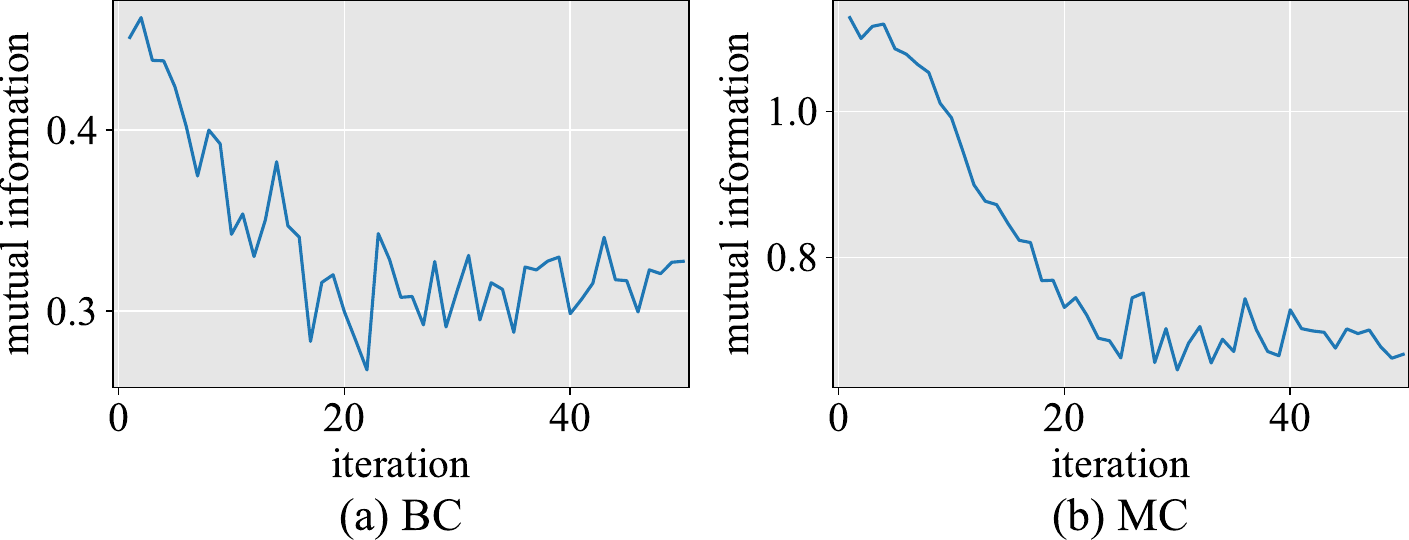}
    \caption{\rev{Mutual information between instance distribution and labels. BC: binary classification; MC: multi-class classification. Higher mutual information implies a higher likelihood of privacy leakage. The mutual information sharply decreases in the first few trees and then starts fluctuating. }}
    \label{fig:mutual}
\end{figure}

We utilize mutual information~\cite{PhysRevE.69.066138} to measure the correlation between labels and the instance distribution on each tree during the training of SecureBoost, to investigate the occurrence of label leakage. Specifically, we trained 10 SecureBoost models, each consisting of 50 trees. Fig.~\ref{fig:mutual} illustrates the average variation of mutual information with respect to iterations (each representing a tree). It indicates that the mutual information is relatively high in the first few iterations of training, suggesting a higher likelihood of privacy leakage during the initial stages of SecureBoost training.

To mitigate the label leakage, we propose a local training stage preceding SecureBoost, where the active party trains $n_l$ decision trees locally. The objective is to reduce the mutual information between the instance distribution and instance labels in the federated SecureBoost model.


The decision trees obtained from the local training stage are stored exclusively on the active party's side, ensuring no information leakage while still participating in the federated prediction process. Once the local training stage is completed, the federated learning stage begins, where all participants collectively train $n_f$ decision trees according to the SecureBoost protocol. These two stages together establish the SecureBoost model with privacy protection.





\begin{figure}[!h]
    \centering \includegraphics[width=0.80\linewidth]{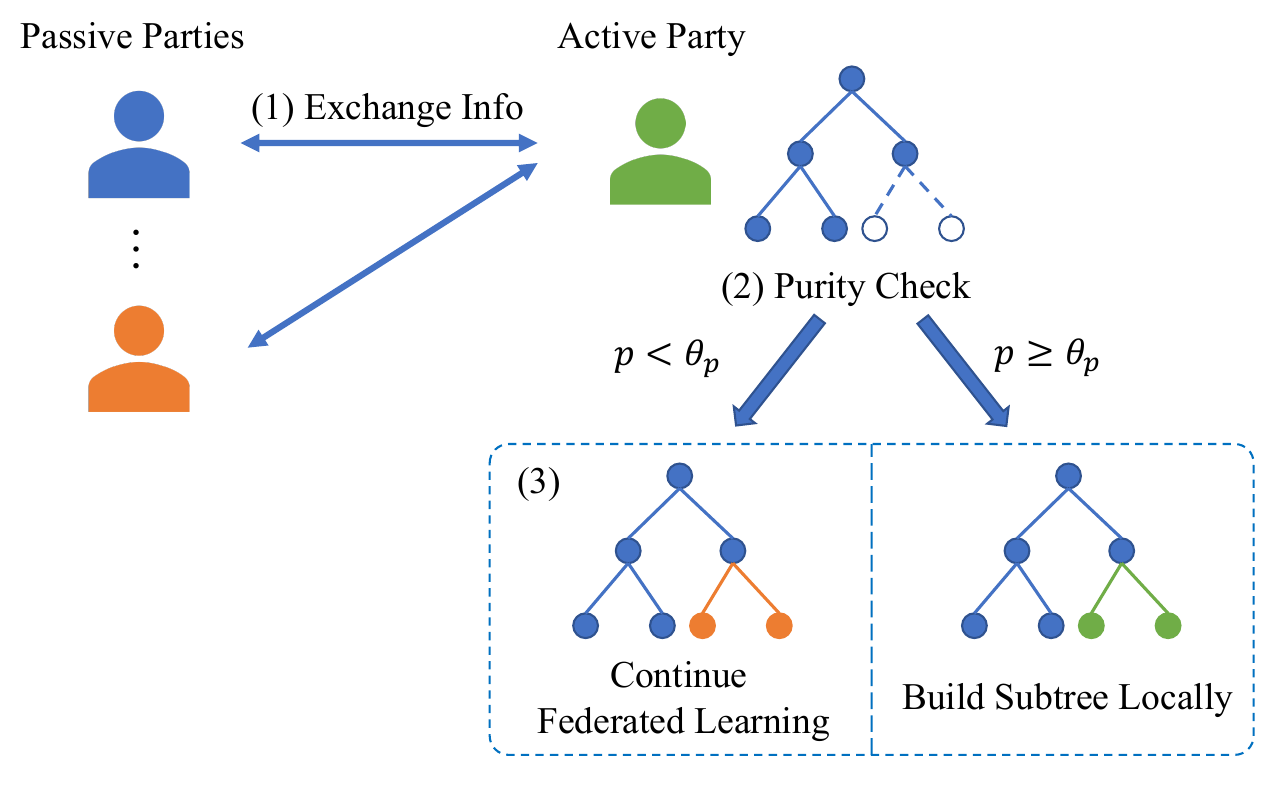}
    \caption{Illustration of the purity threshold method. 
    (1) The active party sends instance distribution and calculates the optimal split point based on statistical data. 
    (2) The active party performs a purity check of the optimal split point. 
    (3) If the purity exceeds $p$ the threshold $\theta_p$, it builds the subtree locally; otherwise, the training continues normally.
    }
    \label{fig:defensemethod}
    \vspace{-2mm}
\end{figure}

\fakeparagraph{Purity Threshold}

The purity threshold mitigates privacy leakage by preventing the sharing of instance distributions of high-purity nodes with passive parties. The active party sends the instance distributions to passive parties, calculates the information gain based on the returned statistical data, and identifies the optimal split point of a node $j$. Subsequently, the active party calculates its purity $p_j$, and if the purity $p_j$ exceeds the pre-specified threshold $\theta_p$, passive parties are excluded from the federated training of the subtree rooted at node $j$, and the active party proceeds to train this subtree locally. Otherwise, all participants continue with the SecureBoost algorithm.

Although the two defense methods can effectively thwart the instance clustering attack, they may introduce utility loss. Therefore, trade-offs need to be made between utility and privacy. We consider their hyperparameters as decision variables when optimizing the Constrained Multi-Objective SecureBoost problem (Sec.~\ref{sec:cmosb}).

\section{Constrained Multi-Objective SecureBoost Algorithm}\label{sec:cmosb}

In this section, we elaborate on our Constrained Multi-Objective SecureBoost algorithm (CMOSB). 

\subsection{Measurements of Objectives}
\label{sec:objectives}

Prior to delving into the algorithmic aspects, we first establish the measurements used to evaluate the three objectives we aim to optimize: utility loss, training cost, and privacy leakage.

\fakeparagraph{Utility Loss}
The utility loss $\epsilon_u$ measures the amount of decrease in utility when certain defense methods are applied to thwart the label inference attack:
\begin{equation}
    \epsilon_{u}=1-U(M, D)
\end{equation}
where $M$, $D$, and $U$ denote the SecureBoost model, the test dataset, and the performance metric, respectively. $U$ is AUC for binary classification and accuracy for multi-class classification. 

\fakeparagraph{Training Cost}
We use training time to measure the training cost of SecureBoost. Since HE operations dominate training time, we estimate the training cost $\epsilon_c$ by summing the time spent on each HE operation:
\begin{equation}
\epsilon_{c}=c_{enc}\times t_{enc}+c_{dec}\times t_{dec}+c_{add}\times t_{add}
\end{equation}
where $c_{enc}$, $c_{dec}$, and $c_{add}$ denote the number of encryption, decryption, and addition operations, respectively, while $t_{enc}$, $t_{dec}$, and $t_{add}$ represent the average time required for each corresponding HE operation.

\fakeparagraph{Privacy Leakage}
We use the accuracy of the instance clustering attack (see Sec. \ref{sec:attack}) to measure the privacy leakage $\epsilon_p$. The attack is evaluated based on a randomly sampled set of instances denoted as $I_{pl}$, where the number of instances belonging to each class is the same. The privacy leakage $\epsilon_p$ can be measured as follows:
\begin{equation}
\epsilon_{p}=\frac{1}{N_{pl}}\sum_{i\in I_{pl}} \mathcal{I}[\hat{y_i}==y_i]
\end{equation}
where $\hat{y_i}$ denotes the label inferred by the instance clustering attack, and $y_i$ denotes the true label, $N_{pl}=|I_{pl}|$ and $\mathcal{I}$ denotes the indicator function. 

\subsection{The CMOSB Algorithm}
\label{sec:mosb}


In this section, we propose the Constrained Multi-Objective SecureBoost algorithm (CMOSB), which aims to obtain approximate Pareto optimal solutions and front. It achieves a balance between the three optimization objectives and provides guidance for hyperparameter selection.


\begin{algorithm}[!h]
	\caption{CMOSB Algorithm}
	\begin{algorithmic}[1]
	\Statex \textbf{Input:} Generations $T$, dataset $D_k$ owned by client $k \in [K]$, constraints $\phi_p, \phi_c$.
    \Statex \textbf{Output:} Pareto optimal solutions and front $\{X_{T}, Y_{T}\}$.
    \State Initialize solutions $\{X_{0}\}$;
    \For{each generation $t$ $=1,2,\cdots,T$} 
        \State Crossover and mutate parent solutions $X_{t-1}$ to produce offspring solutions $P$; 
        \State $R$ $\leftarrow$ Merge $X_{t-1}$ and $P$;
        \State $Y$ $\leftarrow$ SBO$(R, \{D_j\}_{j=1}^K)$ ;
        
        \For{each tuple ($\epsilon_p, \epsilon_c$) in $V$} 
        \State  $\epsilon_{i} = \epsilon_{i} \text{ + } \alpha_i \max\{0, \epsilon_i-\phi_i\}, i \in \{p, c\}$;
        \EndFor
        \State $R^S \leftarrow$ Non-dominated sorting and crowding distance sorting $R$ based on $Y$;
        \State $X_{t}$ $\leftarrow$ Select $N$ high-ranking solutions from $R^S$;
    \EndFor
         \State \Return $\{X_{T}, Y_{T}\}$;
	\end{algorithmic}\label{alg:nsga_fl}
\end{algorithm}

CMOSB is based on NSGA-II and described in Algo.~\ref{alg:nsga_fl}. 
Offspring solutions of the current generation are generated by performing crossover and mutation using solutions found during the previous generation (line 3). All newly generated candidate solutions are evaluated using Algo.~\ref{alg:flo} (line 5). The algorithm adds a constraint to limit the search space of the solutions (lines 6-7). Lines 8-9 sort the candidate solutions and select the top $N$ for the next generation.

\begin{algorithm}[!h]
    \caption{SecureBoost Optimization (SBO)}
	\begin{algorithmic}[1]
    \Statex \textbf{Input:} Solutions X, feature data $D_k$ owned by client $k$.
    \Statex \textbf{Output:} Objective values $Y$ for $X$.
    \For{each solution $x$ $\in$ $X$}
    
    \State{Set hyperparameters $n_l$, $n_f$, $p$ according to $x$; 
    }
    \For{each boosting iteration $i \in 1, \dots, n_l$}
        \State \gray{$\triangleright$ \textit{Local training stage}}
        \State {Active party train decision tree $i$ locally; }
    \EndFor
    \For{each boosting iteration $i \in n_l + 1, \dots, n_l + n_f$}
        \State \gray{$\triangleright$ \textit{Federated learning stage}}
        \State {Active party computes gradient $\left\langle g\right\rangle$, hessian $\left\langle h\right\rangle$; }
        \For{each node $j$ satisfies depth criteria}
            \State {Active party computes purity $p_j$ of node $j$;}
            \If{$p_j < \theta_p$}
            \State {Get instance space $I_{i, j}$ of current node; }
            \State {Split the node using $\text{SF}(I_{i, j}, \left\langle g\right\rangle, \left\langle h\right\rangle)$; }
            \Else
            \State {Active party split the node locally; }
            \EndIf
        \EndFor
        \State {Measure training cost $\epsilon_{e, i}$;}
        \State {Measure privacy leakage $\epsilon_{p, i}$;}
        \State {$\epsilon_{e} \leftarrow \epsilon_{e} + \epsilon_{e, i}$;}
        \State {$\epsilon_{p} \leftarrow \max(\epsilon_{p}, \epsilon_{p, i})$;}
    \EndFor
    \State {Measure utility loss $\epsilon_{u}$;}
    \State $Y \leftarrow Y + (\epsilon_p, \epsilon_c, \epsilon_u)$;
    \EndFor
    \State \Return $Y$;
    \end{algorithmic}\label{alg:flo}
\end{algorithm}

Algo. \ref{alg:flo} aims to measure the utility loss, training cost, and privacy leakage while training a SecureBoost model.
The variable $x$ represents a set of hyperparameters for SecureBoost, and Table~\ref{tab:variable} provides a detailed description of these hyperparameters.
Lines 3-5 correspond to the local training stage mentioned in Sec.~\ref{sec:defense}, while lines 6-19 constitute the federated learning stage, forming the SecureBoost framework.
In line 10, the active party calculates the purity of node $j$, and if it is below the threshold $\theta_p$, all participants will jointly calculate the split point; otherwise, only the active party will calculate it locally.
Training cost and privacy leakage will be calculated during the iteration process, while utility will be calculated after the training is completed.

%% file: sections/5_exp.tex
\section{Experiments}\label{sec:exp}

In this section, we empirically investigate the effectiveness of our proposed attacking method, defense methods, and the hyperparameter optimization algorithm CMOSB.

\subsection{Experimental Settings}

\fakeparagraph{Dataset and setting} 
We conduct experiments on two synthetic datasets and two real-world datasets. The synthetic datasets, generated using the \textit{sklearn} library\footnote{https://scikit-learn.org/stable/}, consist of Synthetic1 for binary classification and Synthetic2 for multi-class classification.
The DefaultCredit
\footnote{https://www.kaggle.com/uciml/default-of-credit-card-clients-dataset} dataset involves predicting whether users can repay their loans on time, while the Sensorless
\footnote{https://archive.ics.uci.edu/dataset/325/} dataset is used for sensorless drive diagnosis.

To create datasets for the federated scenario, each dataset was vertically partitioned into two sub-datasets. We used 2/3 of the data as the training set and the remaining as the test set. Table~\ref{tab:dataset} summarizes these datasets.

We implement all the methods with Python 3.8. The experiments were conducted on two Intel(R) Xeon(R) Platinum 8269CY 3.10GHz CPUs each with 4 cores.

\begin{table}[H]
    \centering  
    \caption{Datasets for evaluation. S: \# of samples; PF: \# of features in passive party; AF: \# of features in active party; C: \# of classes.}  
    \label{tab:dataset}  
    \begin{tabular}{c|c|c|c|c}  
        \hline  
        Name & S & AF & PF & C\\  
        \hline
        \hline
        Synthetic1 & 2,000 & 5 & 5 & 2\\
        \hline
        DefaultCredit & 30,000 & 12 & 13 & 2\\
        \hline
        Synthetic2 & 10,000 & 5 & 5 & 10\\
        \hline
        Sensorless & 58,509 & 12 & 36 & 11\\
        \hline
    \end{tabular}
\end{table}

\fakeparagraph{Baseline} We compared our Constrained Multi-objective SecureBoost (CMOSB) method with the following methods: 


\begin{itemize}

\item Empirical Selection (ES): Empirical Selection represents the default hyperparameters that are typically determined empirically.

\item Grid Search (GS): Grid Search is a traditional hyperparameter optimization method that exhaustively evaluates different combinations for the optimal hyperparameter combination \cite{holly2022evaluation}.

\item Bayesian Optimization (BO): Bayesian Optimization~\cite{DBLP:conf/nips/SnoekLA12} is a classic hyperparameter optimization method for machine learning algorithms. It utilizes Gaussian processes to improve search efficiency. However, BO is only suitable for single-objective optimization. For multi-objective optimization, we leverage BO to optimize each objective separately.

\end{itemize}

\fakeparagraph{Hyperparameter}
We chose three sets of hyperparameters for the ES baseline. The first two sets of hyperparameters are chosen from the default hyperparameters of FATE~\cite{DBLP:journals/jmlr/LiuFCXY21} and VF\textsuperscript{2}Boost~\cite{DBLP:conf/sigmod/FuSYJXT021}, respectively. The third set of hyperparameters is the average of the first two.

To make a fair comparison, the sampling rate is set to $0.8$, and \textit{complete secure} is set to true. 

The \textit{complete secure} corresponds to the local tree defense method, where $n_l$ is set to $1$, indicating that local training of a single tree is performed before federated learning. 

The specific hyperparameter settings are shown in Table~\ref{tab:hp}.

\begin{table}[!h]
    \centering  
    \caption{Default hyperparameters used in the comparison. } 
    \label{tab:hp}  
    \begin{tabular}{c|c|c|c|c|c}  
        \hline  
        Baseline&$n_f$&$d$&$\eta$&$r$&\makecell[c]{complete\\secure}\\  
        \hline
        \hline
        FATE&5&3&0.3&0.8&true\\
        \hline
        VF\textsuperscript{2}Boost&20&7&0.1&0.8&true\\
        \hline
        Average&10&5&0.3&0.8&true\\
        \hline
    \end{tabular}
\end{table}

\begin{table*}[!ht]
    \centering  
    \caption{Hyperparameter variables used in CMOSB algorithm.} 
    \label{tab:variable} 
    \begin{tabular}{c|c|c|c}  
        \hline  
        Variable&Range&Chromosome Type&Description \\  
        \hline
        \hline
        $n_f$&$[1, 16]$&Binary&The number of iterations for boosting in federated learning\\
        \hline
        $n_l$&$[1, 16]$&Binary&The number of iterations for boosting locally\\
        \hline
        $d$&$[1, 8]$&Binary&Maximum depth of each decision tree \\
        \hline
        $r$&$[0.1, 1.0]$&Real-value&Subsample ratio of the training instances\\
        \hline
        $\theta_p$&$[0.1, 1.0]$&Real-value&The purity threshold to stop the federated learning\\
        \hline
        $\eta$&$[0.01, 0.3]$&Real-value&Step size shrinkage used in update to prevents overfitting\\
        \hline
    \end{tabular}
\end{table*}

\fakeparagraph{CMOSB Setup} 
The proposed CMOSB algorithm is based on NSGA-II. Thus, we follow the setup proposed in literature~\cite{DBLP:journals/tnn/ZhuJ20}. 
We apply a single-point crossover for binary chromosomes with a probability of 0.9 and a bit-flip mutation with a probability of 0.1. 
We apply a simulated binary crossover (SBX)~\cite{DBLP:journals/compsys/DebK95} for real-valued chromosome with a probability of 0.9 and $n_c$ = 2, and a polynomial mutation with a probability of 0.1 and $n_m$ = 20, where $n_c$ and $n_m$ denote spread factor distribution indices for crossover and mutation, respectively.
The number of generations for CMOSB is set to 40.






\subsection{Effectiveness of Defense Methods}
\label{sec:exp-def}

In this section, we investigate the efficacy of our proposed defense methods in mitigating privacy leakage. The experiments are conducted on Synthetic1 dataset, with the default hyperparameter set as follows: $n=20$, $d=7$, $\eta=0.1$, $r=0.8$.

\begin{figure}[!h]
    \centerline{
        \begin{minipage}[t]{0.6\linewidth}
            \centering
            \includegraphics[width=\textwidth]{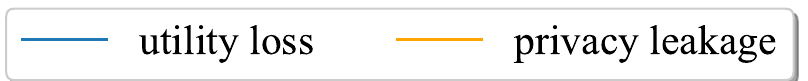}
        \end{minipage}%
    }
    \vspace{2mm}
    \begin{minipage}[t]{0.99\linewidth}
        \centering
        \includegraphics[width=\textwidth]{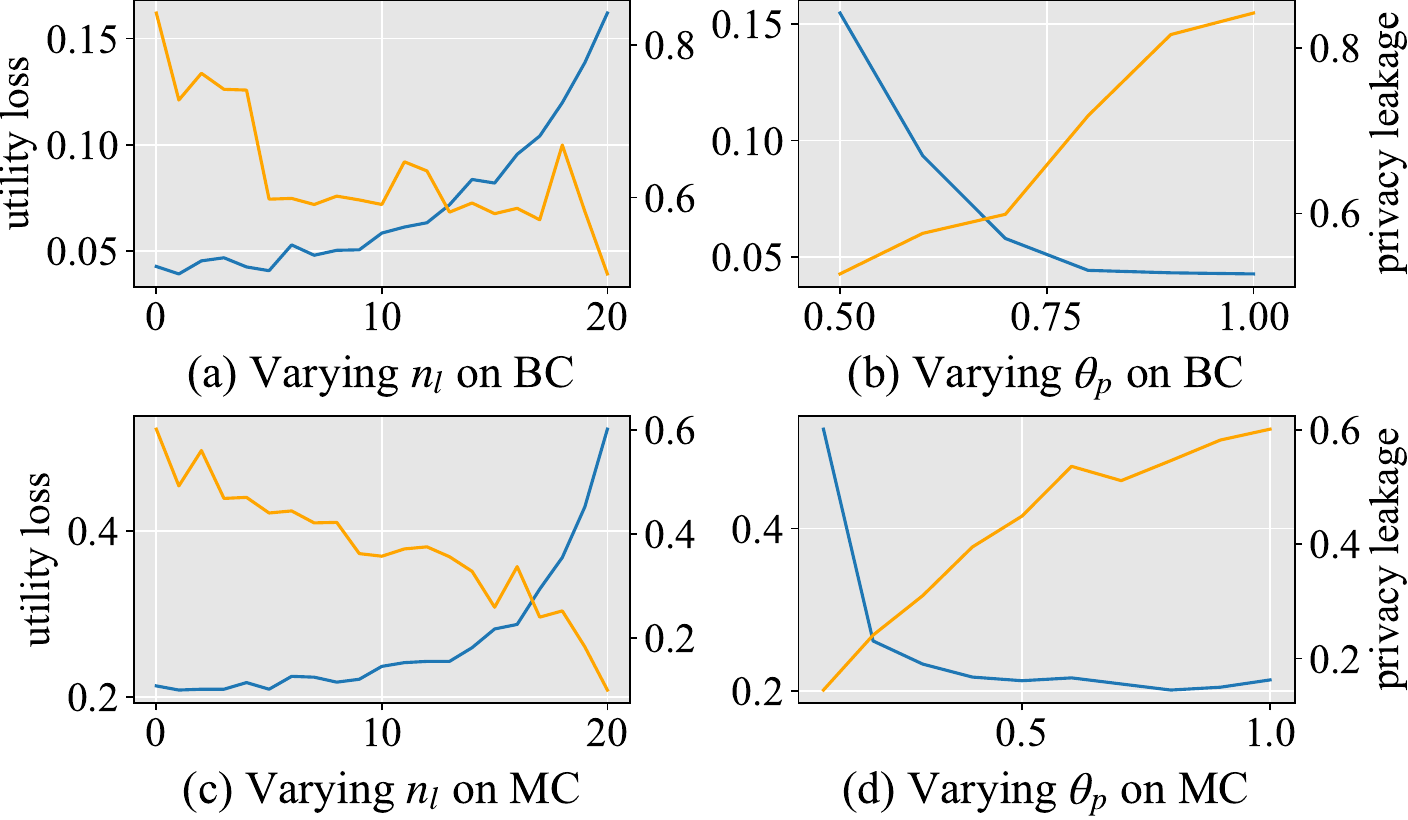}
    \end{minipage}%
    \caption{
    \rev{
    Effectiveness of Defense Methods.
    The yellow line represents privacy leakage, where lower values indicate a more secure model. The blue line represents utility loss, where lower values indicate better model performance. The first and second columns are experiments on Local Trees and Purity Threshold, respectively.
    BC: binary classification; MC: multi-class classification. Reducing $p$ or increasing $n_l$ can decrease privacy leakage while sacrificing utility.
    }
    }
    \label{fig:var}
\end{figure}

Fig.~\ref{fig:var} shows that both methods can trade-off model performance for privacy protection. 
The lack of protection on instance distribution resulted in up to 84\% of instance label leakage. 
As shown in Fig.~\ref{fig:var} (a), training 10 decision trees locally reduced privacy leakage risk by 25.1\% while sacrificing only 1.6\% of utility.
The impact of the purity threshold is illustrated in Fig.~\ref{fig:var}(b), where we observe a decrease in privacy leakage as the active party constructs more nodes locally.

\subsection{Optimal Tradeoffs between Utility, Efficiency, and Privacy achieved by CMOSB}

In this section, we use CMOSB (Algo.~\ref{alg:nsga_fl}) without constraints to find the Pareto optimal solutions of hyperparameters for SecureBoost. Each solution is a set of hyperparameters that achieves an optimal tradeoff between utility loss, training cost, and privacy leakage.

We summarize the variables and their descriptions involved in multi-objective optimization in Table~\ref{tab:variable}. Among the variables, the definitions of $n_f$, $d$, $r$, and $\eta$ are the same as those used in SecureBoost~\cite{DBLP:journals/expert/ChengFJLCPY21}. $n_l$ refers to the number of iterations for boosting locally, and $\theta_p$ refers to the threshold of purity. For binary classification problems, we further constrain the range of $\theta_p$ to $[0.7, 1.0]$ to improve search efficiency.



\begin{figure}[H]
    \centerline{
        \begin{minipage}[t]{0.8\linewidth}
            \centering
            \includegraphics[width=\textwidth]{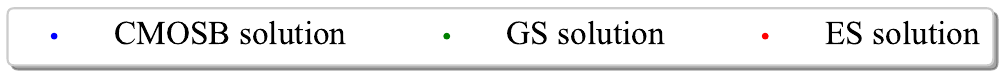}
        \end{minipage}%
    }
    \begin{minipage}[t]{0.49\linewidth}
        \centering
        \includegraphics[width=\textwidth]{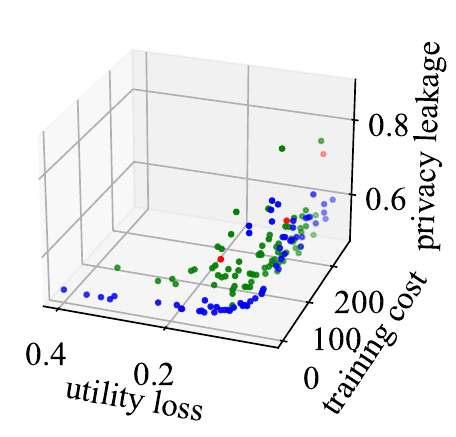}
        \centerline{(a) Synthetic1}
    \end{minipage}%
    \begin{minipage}[t]{0.49\linewidth}
        \centering
\includegraphics[width=\textwidth]{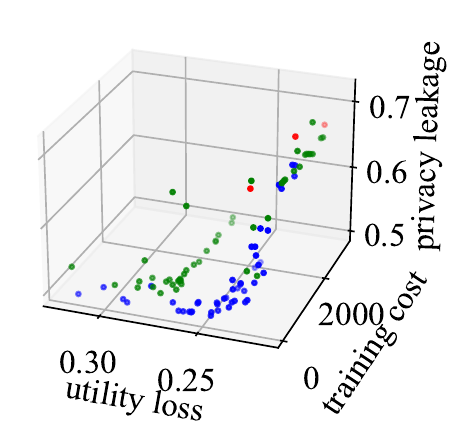}
        \centerline{(b) DefaultCredit}
    \end{minipage}
    \begin{minipage}[t]{0.49\linewidth}
        \centering
        \includegraphics[width=\textwidth]{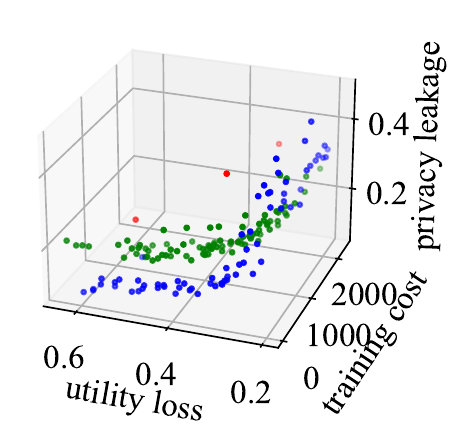}
        \centerline{(c) Synthetic2}
    \end{minipage}%
    \begin{minipage}[t]{0.49\linewidth}
        \centering
        \includegraphics[width=\textwidth]{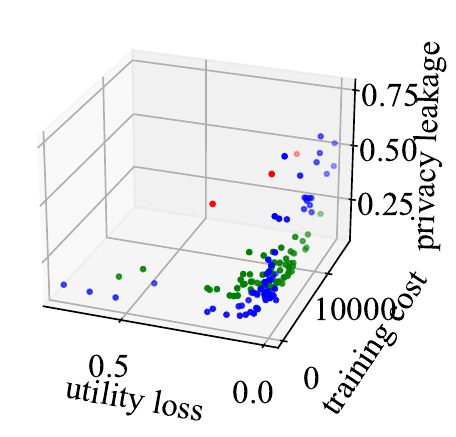}
        \centerline{(d) Sensorless}
    \end{minipage}
    \caption{
    Comparison between 3D Pareto fronts of CMOSB and GS (Grid Search). The blue, green, and red dots represent solutions obtained by CMOSB, GS, and ES (Empirical Selection), respectively. Solutions closer to the origin are better. 
    }
    \label{fig:exp-3d-gs}
\end{figure}

\begin{figure}[!h]
    \centerline{
        \begin{minipage}[t]{0.8\linewidth}
            \centering
            \includegraphics[width=\textwidth]{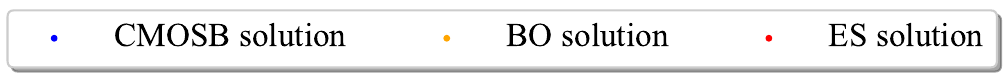}
        \end{minipage}%
    }
    \begin{minipage}[t]{0.49\linewidth}
        \centering
        \includegraphics[width=\textwidth]{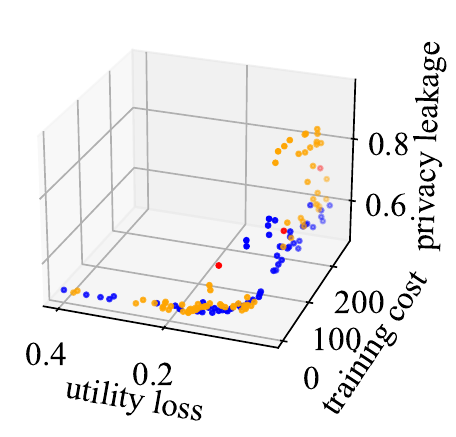}
        \centerline{(a) Synthetic1}
    \end{minipage}%
    \begin{minipage}[t]{0.49\linewidth}
        \centering
\includegraphics[width=\textwidth]{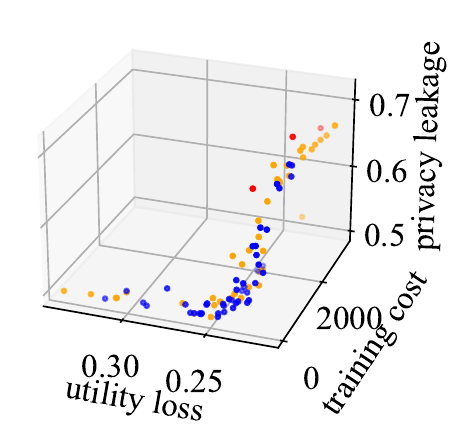}
        \centerline{(b) DefaultCredit}
    \end{minipage}
    \begin{minipage}[t]{0.49\linewidth}
        \centering
        \includegraphics[width=\textwidth]{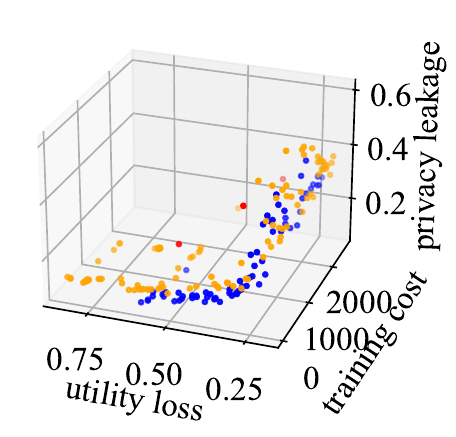}
        \centerline{(c) Synthetic2}
    \end{minipage}%
    \begin{minipage}[t]{0.49\linewidth}
        \centering
        \includegraphics[width=\textwidth]{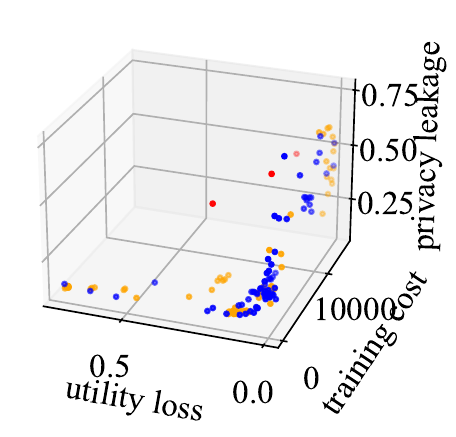}
        \centerline{(d) Sensorless}
    \end{minipage}
    \caption{
    Comparison between 3D Pareto fronts of CMOSB and BO (Bayesian Optimization). The blue, orange, and red dots represent solutions of CMOSB, BO, and ES, respectively. Solutions closer to the origin are better. 
    }
    \label{fig:exp-3d-bayes}
\end{figure}

We first present the comparison between CMOSB and the baselines through 3D plots illustrated in Fig.~\ref{fig:exp-3d-gs} and Fig.~\ref{fig:exp-3d-bayes} on the four datasets (see Table \ref{tab:dataset}).

Fig.~\ref{fig:exp-3d-gs} illustrates the comparison between CMOSB and Grid Search. The red, green, and blue dots represent solutions obtained from Empirical Selection (ES), Grid Search (GS), and CMOSB, respectively. It can be observed that the solutions found by CMOSB dominate those found by ES and GS, indicating that CMOSB achieves a better Pareto front (closer to the origin).

Fig.~\ref{fig:exp-3d-bayes} illustrates the comparison between CMOSB and Bayesian optimization. The red, orange, and blue dots represent solutions obtained from Empirical Selection (ES), Bayesian Optimization (BO), and CMOSB, respectively. Similar to Fig.~\ref{fig:exp-3d-gs}, CMOSB outperforms BO and ES overall, achieving a better Pareto front.

\begin{figure*}[!h]
    \centerline{
        \begin{minipage}[t]{0.99\linewidth}
            \centering
            \includegraphics[width=\textwidth]{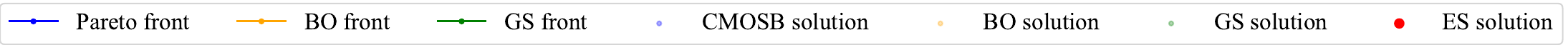}
        \end{minipage}%
    }
    \vspace{8pt}
    \begin{minipage}[t]{0.99\linewidth}
        \centering
        \includegraphics[width=\textwidth]{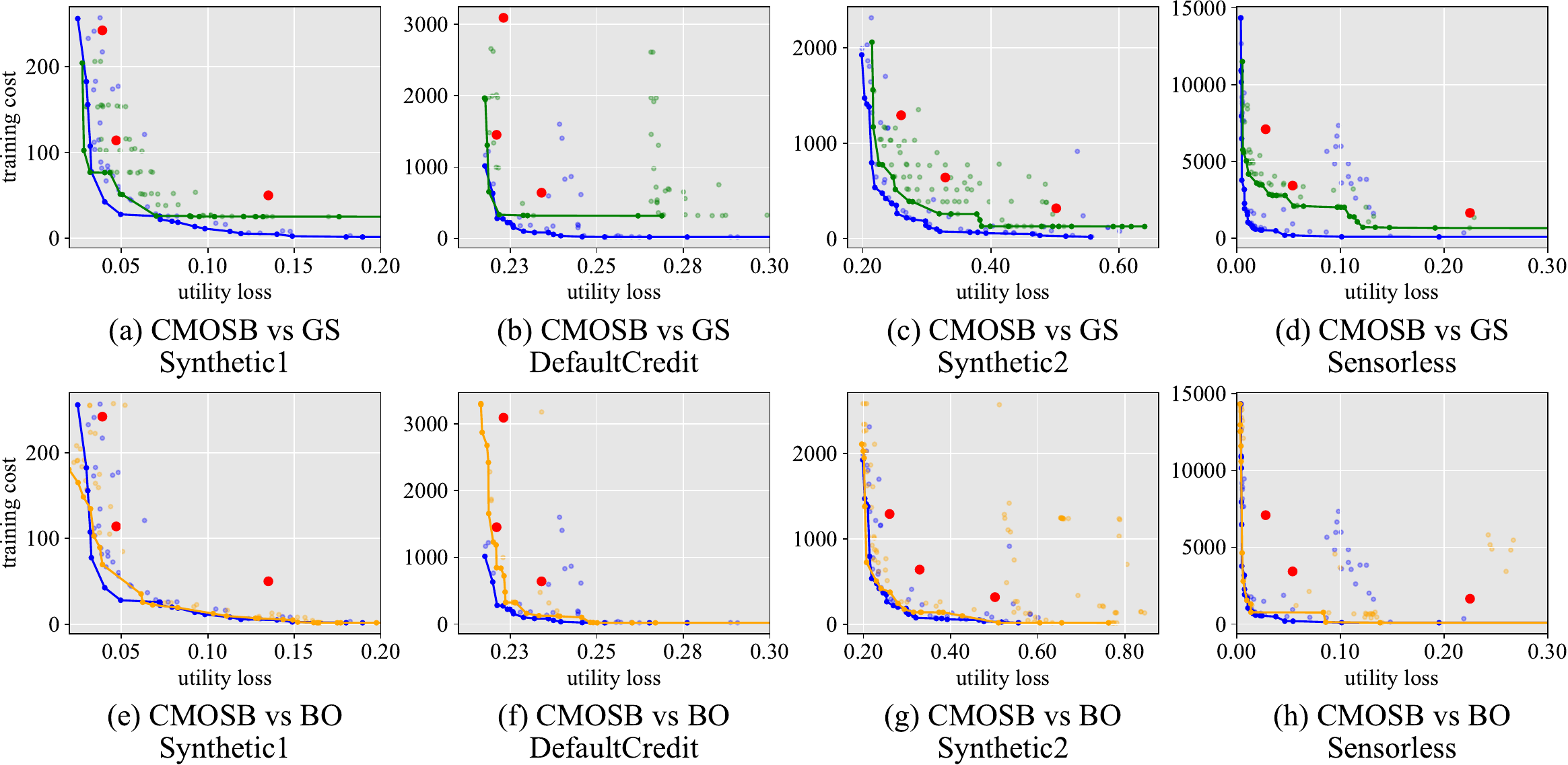}
    \end{minipage}
    \caption{
    Comparing the Pareto front of CMOSB with baseline methods in terms of the tradeoff between Training Cost and Utility Loss. The first two columns are for two binary classification tasks, Synthetic1 and Credit2, while the last two columns are for multi-class classification tasks, Synthetic2, and Sensorless. In each sub-figure, solutions closer to the bottom-left corner are considered better.
    }
    \label{fig:cmp-tc}
    \vspace{10pt}
\end{figure*}

\begin{figure*}[!h]
    \centerline{
        \begin{minipage}[t]{0.99\linewidth}
            \centering
            \includegraphics[width=\textwidth]{figure/exp/legend/legend-pf.pdf}
        \end{minipage}%
    }
    \vspace{8pt}
    \begin{minipage}[t]{0.99\linewidth}
        \centering
        \includegraphics[width=\textwidth]{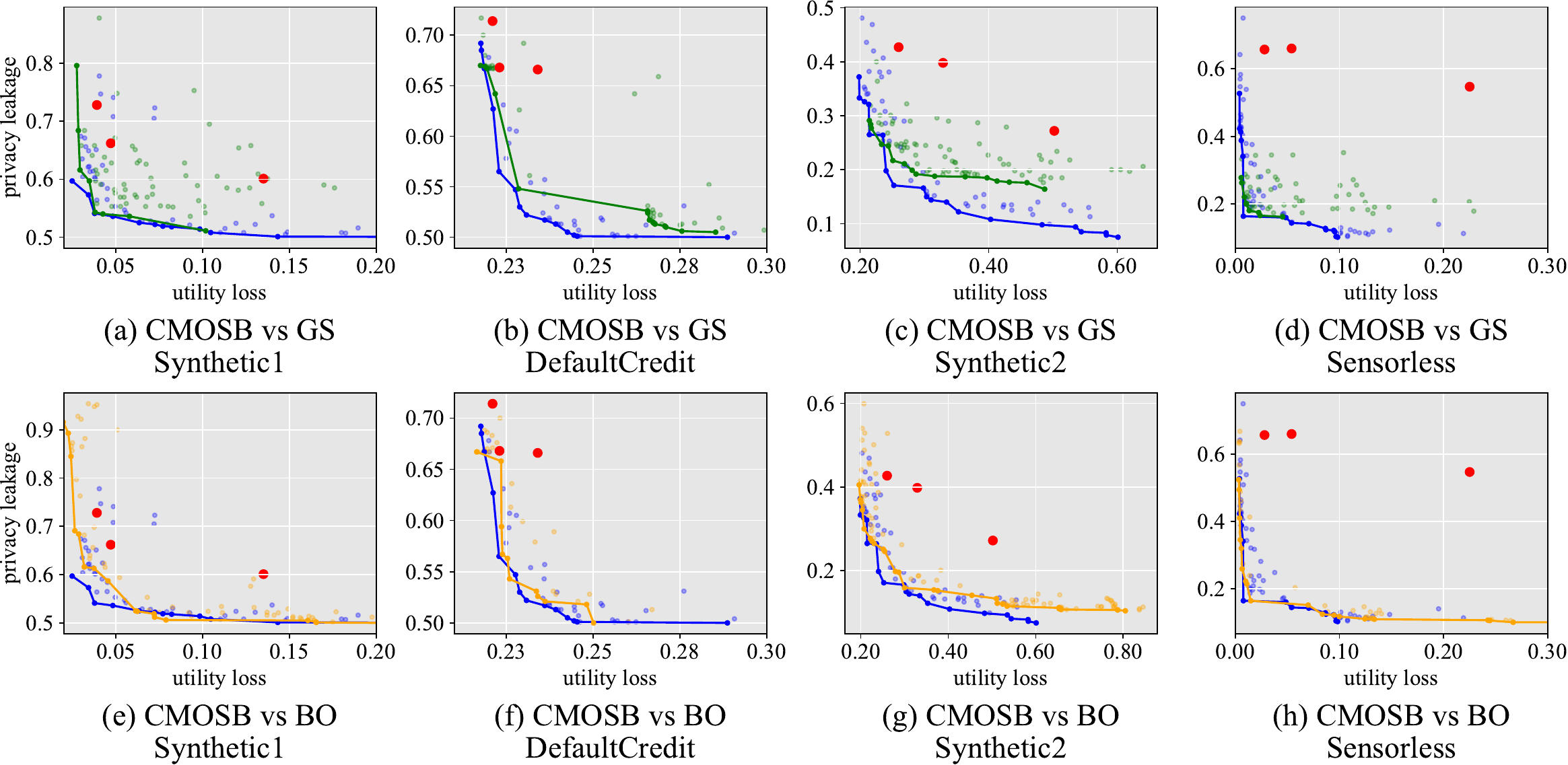}
    \end{minipage}
    \caption{
    Comparing the Pareto front of our proposed CMOSB with baseline methods in terms of the tradeoff between Privacy Leakage and Utility Loss. The first two columns are for two binary classification tasks, Synthetic1 and Credit2, while the last two columns are for multi-class classification tasks, Synthetic2, and Sensorless. In each sub-figure, solutions closer to the bottom-left corner are considered better.
    }
    \label{fig:cmp-pl}
\end{figure*}

To provide a more intuitive observation of the experimental results, we project each 3D plot onto two 2D plots: one for training cost vs. utility loss and another for privacy leakage vs. utility loss, as shown in Fig.~\ref{fig:cmp-tc} and Fig.~\ref{fig:cmp-pl}, respectively. 
In Fig.~\ref{fig:cmp-tc} and Fig.~\ref{fig:cmp-pl},
the blue, green, orange, and red dots represent CMOSB, GS~(Grid Search), BO~(Bayesian Optimization), and ES~(Empirical Selection), respectively. 
Each dot represents a solution, and each line represents a Pareto front.


For the trade-off between training cost and utility loss (see Fig.~\ref{fig:cmp-tc}), all four datasets demonstrate that the training cost increases as the utility loss decreases. CMOSB (in blue) can find a better Pareto front than GS (in green), BO (in orange), and ES (in red). More specifically, BO (in orange) can find a set of decent solutions but is slightly inferior to CMOSB. This is because BO optimizes one objective at a time and does not consider all three objectives as a whole for optimization. 
The solutions found by GS are noticeably inferior to CMOSB, as GS only heuristically searches for solutions.
The solutions achieved by ES are farther away from the lower-left corner, indicating that ES performs poorly in simultaneously optimizing utility loss, training cost, and privacy leakage. This suggests that the default hyperparameters set by FATE~\cite{DBLP:journals/jmlr/LiuFCXY21} and VF\textsuperscript{2}Boost~\cite{DBLP:conf/sigmod/FuSYJXT021} were not suitable for multi-objective SecureBoost.

For the trade-off between privacy leakage and utility loss (see Fig.~\ref{fig:cmp-pl}), all four datasets demonstrate that as the utility loss decreases, the privacy leakage increases. CMOSB outperforms all baseline methods in terms of the Pareto front. BO performs slightly worse than CMOSB because it is a single-objective optimization algorithm and thus cannot simultaneously optimize privacy leakage, utility loss, and efficiency.
GS demonstrates inferior search results, as applying grid search to optimize continuous hyperparameters is often challenging. More specifically, the privacy protection threshold $p$, which significantly impacts privacy leakage, is a continuous variable, and the performance of grid search is limited in this case. ES overall performs quite badly in preserving label privacy (the worst among all methods) because FATE and VF\textsuperscript{2}Boost were not well-prepared for defending against ICA.

\begin{table}[!h]
    \centering  
    \caption{Hypervolume Comparison between our CMOSB and baseline methods, including ES (Empirical Selection), GS (Grid Search), and BO(Bayesian Optimization) .} 
    \label{tab:cmp-hv}  
    \begin{tabular}{c|c|c|c|c}  
        \hline  
        Baseline&Synthetic1&Credit&Synthetic2&Sensorless\\  
        \hline
        ES&0.516&0.094&0.368&0.134\\
        \hline
        GS&0.826&0.746&0.706&0.822\\
        \hline
        BO&0.902&0.900&0.805&0.964\\
        \hline
        CMOSB&\textbf{0.920}&\textbf{0.920}&\textbf{0.879}&\textbf{0.968}\\
        \hline
    \end{tabular}
\end{table}

We also compare the hypervolume of CMOSB with those of baseline methods (see Table~\ref{tab:cmp-hv}) for a clear performance comparison between the four hyperparameter tuning methods. We normalize the three objectives for each dataset to ensure a fair comparison. The performance of ES varies widely across the four datasets due to its empirical selection of default hyperparameters for SecureBoost.
While obtaining relatively stable results by searching a large number of hyperparameters, GS suffers from a heuristically pre-specified discrete hyperparameter search space, which is not well-suited for our proposed two defense methods, as they are sensitive to their defense parameters (i.e., $n_l$ and $p$). BO performs better than ES and GS but worse than CMOSB because BO optimizes each objective separately and does not consider all objectives as a whole for optimization.





\subsection{Multi-Objective SecureBoost under Constraints}

In real-world VFL scenarios, participants typically have specific requirements or constraints on objectives. Hence, we apply the CMOSB algorithm (with constraints) to ensure that the Pareto optimal solutions found satisfy the constraints as much as possible. Adding constraints focuses the search space of the CMOSB algorithm on the feasible region, which improves the efficiency of finding better Pareto optimal solutions.


We conduct this experiment on Synthetic1 and constrain the training cost and privacy leakage to be below 100 seconds and 0.6, respectively. We set the penalty coefficient to 20 (see lines 6-7 of Algorithm \ref{alg:nsga_fl}).

\begin{figure}[!h]
    \begin{minipage}[t]{0.49\linewidth}
        \centering
        \includegraphics[width=\textwidth]{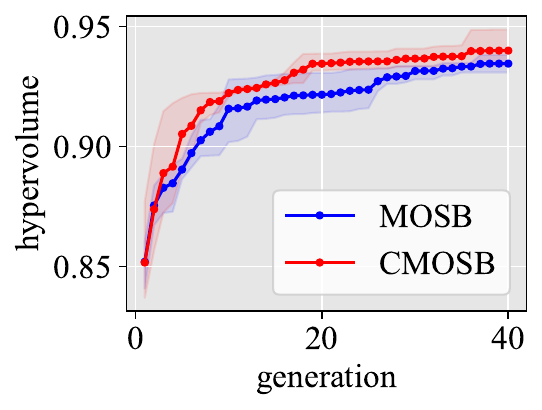}
        \centerline{(a) PL Constraint}
    \end{minipage}%
    \begin{minipage}[t]{0.49\linewidth}
        \centering
        \includegraphics[width=\textwidth]{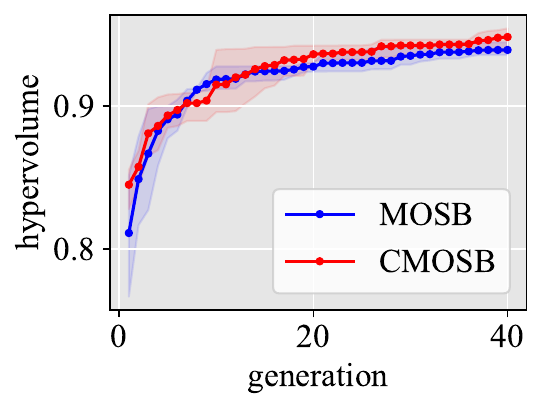}
        \centerline{(b) TC Constraint}
    \end{minipage}%
    \caption{
    Comparison of hypervolume between MOSB and CMOSB algorithms. 
    TC: training cost; PL: privacy leakage. 
    The red line represents CMOSB, and the blue line represents MOSB.
    A higher hypervolume implies that better solutions can be found. 
    }
    \label{fig:constraint-hv}
\end{figure}

Fig.~\ref{fig:constraint-hv} illustrates the hypervolume comparison between CMOSB and MOSB when applying constraints to training cost and privacy leakage, respectively. For the privacy leakage constraint, CMOSB grows more rapidly in the first few generations and surpasses MOSB in the final generation, indicating that it can effectively find better Pareto solutions. For the training cost constraint, CMOSB also surpasses MOSB after the 10th generation and maintains a lead until the last generation.


\begin{figure}[!h]
    \centerline{
        \begin{minipage}[t]{0.99\linewidth}
            \centering
            \includegraphics[width=\textwidth]{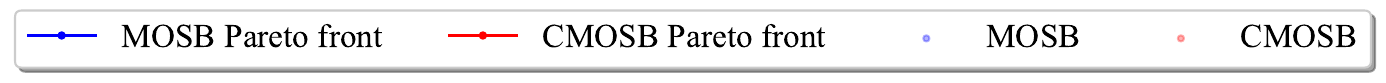}
        \end{minipage}%
    }
    \vspace{2mm}
    \begin{minipage}[t]{0.49\linewidth}
        \centering
        \includegraphics[width=\textwidth]{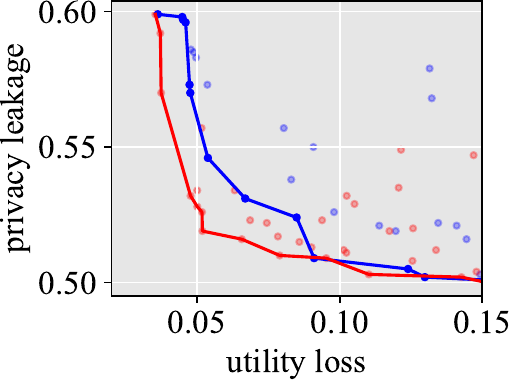}
        \centerline{(a) PL vs UL}
    \end{minipage}%
    \begin{minipage}[t]{0.51\linewidth}
        \centering
        \includegraphics[width=\textwidth]{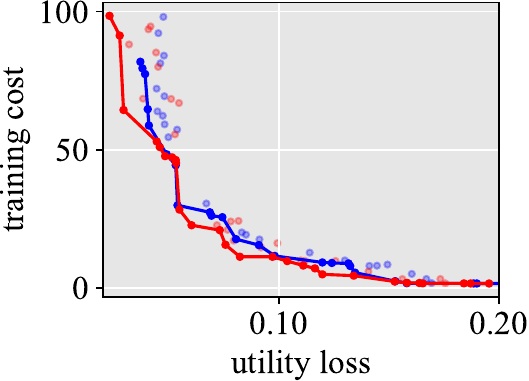}
        \centerline{(b) TC vs UL}
    \end{minipage}
    \caption{
    Comparison of Pareto front between MOSB and CMOSB algorithm. 
    UL: utility loss; TC: training cost; PL: privacy leakage. 
    We add a constraint on PL in Figure(a), and add a constraint on TC in Figure(b). 
    The red line represents CMOSB, and the blue line represents MOSB. 
    The solutions located closer to the bottom left corner of the graph are considered better.
    }
    \label{fig:constraint-2d-pareto}
\end{figure}

We further compare the Pareto fronts (at the 40th generation) obtained by the CMOSB and MOSB algorithms. Fig.~\ref{fig:constraint-2d-pareto}(a) demonstrates the effect of adding a privacy leakage constraint, showing a reduction of approximately 3\% in privacy leakage of the Pareto solutions at the same level of utility loss. The Pareto front found by CMOSB in Fig.~\ref{fig:constraint-2d-pareto}(b) is also superior to that found by MOSB, especially for solutions with lower utility loss.

%% file: sections/6_conclusion.tex
\section{Conclusion}
\label{sec:conclusion}

In this paper, we address two main limitations of the SecureBoost algorithm: privacy leakage and hyperparameter optimization. We first propose the \textit{instance clustering attack (ICA)} that can infer the labels of active party and then we develop two defense methods that can thwart ICA. Next, we propose the Constrained Multi-Objective SecureBoost (CMOSB) algorithm, which identifies Pareto optimal solutions of hyperparameters for SecureBoost by simultaneously minimizing utility loss, training cost, and privacy leakage. 
We conduct experiments on four datasets to validate the effectiveness of the proposed ICA and the corresponding defense methods. Furthermore, experimental results demonstrate that the Pareto optimal solutions of hyperparameters found by CMOSB outperform those obtained by the baselines in terms of the Pareto front.